\documentclass{article}
\pdfoutput=1
\PassOptionsToPackage{numbers, compress}{natbib}

\usepackage[preprint]{nips_2018}

\usepackage[utf8]{inputenc} 
\usepackage[T1]{fontenc}    
\usepackage{hyperref}       
\usepackage{url}            
\usepackage{booktabs}       
\usepackage{amsfonts}       
\usepackage{nicefrac}       
\usepackage{microtype}      
\usepackage{wrapfig}

\usepackage{graphicx}
\usepackage{footnote}
\usepackage{threeparttable}
\usepackage{paralist}
\usepackage{mathtools}
\usepackage{color}
\usepackage{multibib}
\usepackage[toc,page]{appendix}
\usepackage{subcaption}
\usepackage{placeins}

\newcites{app}{Appendix References}

\DeclarePairedDelimiterX{\infdivx}[2]{(}{)}{%
  #1\delimsize\|#2%
}

\title{Controllable Semantic Image Inpainting}

\author{
  Jin Xu \\
  University of Oxford\\
  \texttt{aaron.jin.xu@gmail.com} \\
  \And
  Yee Whye Teh \\
  University of Oxford\\
  DeepMind\\
  \texttt{y.w.teh@stats.ox.ac.uk} \\
}

\begin{document}

\maketitle

\begin{abstract}
  We develop a method for user-controllable semantic image inpainting: Given an arbitrary set of observed pixels, the unobserved pixels can be imputed in a user-controllable range of possibilities, each of which is semantically coherent and locally consistent with the observed pixels. We achieve this using a deep generative model bringing together: an encoder which can encode an arbitrary set of observed pixels, latent variables which are trained to represent disentangled factors of variations, and a  bidirectional PixelCNN model. We experimentally demonstrate that our method can generate plausible inpainting results matching the user-specified semantics, but is still coherent with observed pixels. We justify our choices of architecture and training regime through more experiments.
\end{abstract}

\section{Introduction}

\begin{wrapfigure}{r}{.45\textwidth}
\vspace*{-2em}
  \centering
    \includegraphics[width=1.0\linewidth]{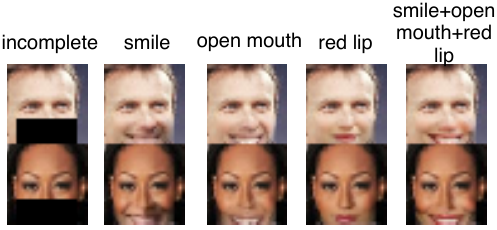}
\caption{Controllable semantic inpainting.}
\vspace*{-1em}
\label{fig:CSI-demo}
\end{wrapfigure}

Semantic image inpainting attempts to infer missing regions of an image based on high-level understandings of the image semantics. It is more challenging than classical inpainting problems such as restoration of corrupted images or removal of selected objects where it is still possible to only use local details and texture \citep{shen2002mathematical,efros1999texture} or prior knowledge about images \citep{he2012statistics,hu2013fast,huang2014image,ulyanov2017deep}. Intuitively, the semantic inpainting algorithm should be able to know the common structure of human faces or understand that it is more appropriate to put a bird rather than a lion on a tree branch. 
Our research is mainly motivated by the limitation of existing semantic inpainting methods that image inpainting results are completely out of users' control. Existing methods either only give deterministic predictions \citep{pathak2016context,yu2018generative,iizuka2017globally}, or even though diversified samples can be generated \citep{yeh2017semantic,li2017generative}, the generating process does not provide us with any means to interact with it and have control over the generated content to some extent. However, when users work on image editing, they very often want to specify their requirements, rather than accepting whatever they get from the algorithm. Therefore, we focus on developing a controllable semantic inpainting method with an interface for manipulating high-level semantics of the inpaintings (Figure \ref{fig:CSI-demo}).
In addition, previous learning-based methods usually involves some form of adversarial losses \citep{goodfellow2014generative} to ensure the inpaintings are plausible \citep{pathak2016context,yu2018generative,iizuka2017globally,yeh2017semantic,li2017generative}. Moreover, post-processing such as Poisson blending \citep{perez2003poisson} is sometimes necessary to ensure local harmony near the boundary of the missing regions \citep{yeh2017semantic}. The proposed method can provide plausible and coherent image completion without the two.

We address the controllable semantic inpainting problem using a deep generative model bringing together several components: an encoder which transforms partially observed images to high-level semantic representations of complete images, latent variables which are trained to be disentangled and interpretable in order to serve as an interface for user control, a new bidirectional PixelCNN model, which is used to extract high-level semantics from the latent variables, and low-level details from surrounding pixels, and in the end, outputs a conditional distribution of the unobserved pixels given the observed ones.

Our contributions can be summarised as follows: 
\begin{inparaenum}[1)]
\item We propose a user-controllable semantic inpainting method, where the generated content matches high-level semantics set by the user, and is both semantically and locally coherent with the observed pixels without post-processing. 
\item We propose a bidirectional PixelCNN model, which is an autoregressive semantic inpainting method in its own right. However, it is only used to capture low-level details and texture in the observed pixels in our proposed method.
\item We analyse the difficulty of designing training objectives under the variational autoencoder framework \citep{kingma2013auto,rezende2014stochastic} when an expressive decoder and disentangled latent representations are both wanted at the same time. Effects of possible objective candidates are explored through experiments.
\end{inparaenum}

\section{A Controllable Semantic Inpainting (CSI) Method} \label{sec:a_controllable_semantic_inpainting_method}

\begin{figure}
  \centering
    \includegraphics[width=1\linewidth]{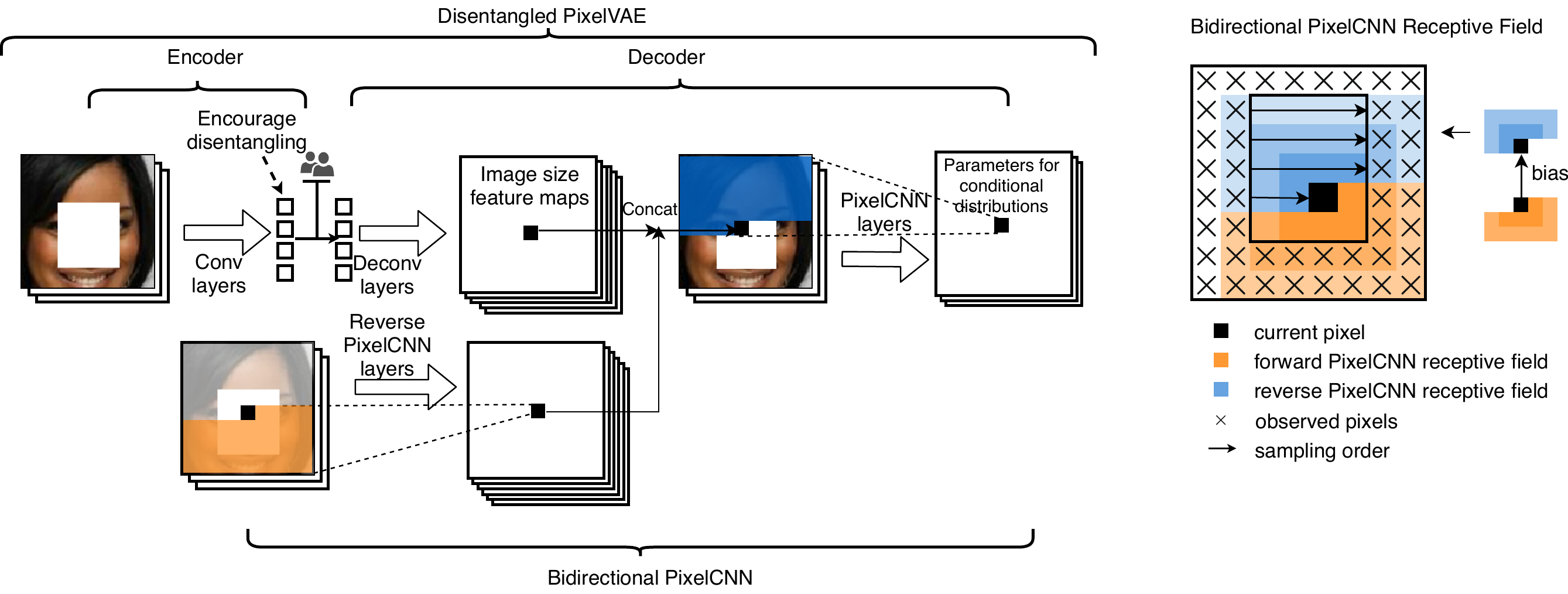}
\caption{Model architecture. Context information is processed using two streams: the disentangled PixelVAE stream extracts high-level semantics, while the bidirectional PixelCNN stream fills in low-level details and textures. Users can manipulate inpainting semantics without sacrificing fine details and textures by varying the latent variables.}
\label{fig:model_architecture}
\end{figure}

Semantic inpainting is the process of imputing the unobserved pixels (called the \emph{target}) in an image given the observed ones (called the \emph{context}), in a way that is coherent at two distinct levels: a high-level coherence based on a semantic understanding of the objects and scenes in the context, and a low-level coherence in terms of local details and textures.  Controllable semantic inpainting (CSI) goes a step further and gives the user an easy way to control, at the semantic level, the inpainting process.
In this paper, we develop a model for controllable semantic inpainting that addresses both levels of coherence, by bringing together a deep generative latent variable model to model the high-level semantics, and a new bidirectional PixelCNN model to handle the low-level details. See Figure \ref{fig:model_architecture} for an overview of the model. The rest of the section gives an overview of and motivation for both parts of the model, while the next two sections describe them in more detail.

We implement user control for semantic inpainting via manipulating the values of latent variables in the generative process. In order that the manipulation is intuitive, we desire latent variables that correspond to a disentangled and more interpretable semantic representation of the image \citep{chen2016infogan,higgins2016beta,chen2018isolating,kim2018disentangling}. To achieve this, we use a variational auto-encoder (VAE) \citep{kingma2013auto,rezende2014stochastic} type model trained using an objective that encourages disentangling \citep{higgins2016beta,chen2018isolating,kim2018disentangling,zhao2017infovae}.  Section \ref{sec:disentangled_pixelvae} provides a discussion of different objectives that have been recently proposed for this purpose (we decided on the InfoVAE-MMD objective \citep{zhao2017infovae}). The network architecture of latent variable model follows that of the PixelVAE \citep{gulrajani2016pixelvae}: it has a convolutional encoder which transforms spatial image features into a semantic space, and a decoder which has deconvolutional layers, which transform the semantic space into spatial feature maps, followed by PixelCNN \citep{oord2016pixel,van2016conditional} module, which transforms the feature maps back into an image. Section \ref{sec:disentangled_pixelvae} provides more details on the disentangled PixelVAE.

The PixelCNN module model the decoded image using a powerful autoregressive model which captures the low-level details well \citep{oord2016pixel,van2016conditional,gulrajani2016pixelvae}. Each pixel is modelled using a distribution conditional on the previous pixels in the image in a top-to-bottom, left-to-right raster scan order. This pixel ordering implies that the PixelCNN layers can only directly model dependence on the context pixels on the top and left of each pixel (as well as the previously generated target pixels) (blue section in Figure \ref{fig:model_architecture}), while dependence on context pixels in the bottom and right are modelled only indirectly through the latent variables, which cannot capture low-level details. This can lead to glaring discontinuities on the bottom and right edges of target regions. We propose to remedy this by using a second PixelCNN which operates in the reverse (bottom-to-top and right-to-left) direction, and which is applied only to the context pixels (the target pixels are masked) (orange section in Figure). The outputs of both the forward and reverse PixelCNNs are finally combined together to form the distribution of each unobserved pixel, conditioned on all context pixels and on previous target pixels in the forward order (thus capturing all relevant low-level details), and on the latent variables through the deconvolutional layers (thus capturing high-level semantics). We call this architecture the bidirectional PixelCNN, and describe it in more detail in Section \ref{sec:bidirectional_pixecnn}. Note that the forward PixelCNN is a module in both the disentangled PixelVAE and the bidirectional PixelCNN components of the model.

Figure \ref{fig:model_architecture} summarises the model architecture. Context information is fed into the system as inputs for both the disentangled PixelVAE, and the bidirectional PixelCNN. It is our intention to learn semantics using the disentangled PixelVAE and details from the bidirectional PixelCNN. However, the bidirectional PixelCNN is an effective semantic inpainting method on its own right (just not a controllable one), and we found no way to prevent it from learning high-level semantics, if the model is trained end-to-end. This prevents the disentangled PixelVAE from learning a good high-level semantic representation. We find that the problem can be effectively solved by two-stage training. During the first stage, only the disentangled PixelVAE part is trained with an objective which targets recovering a complete image from the context, while disentangling the latent variables. During the second stage, the bidirectional PixelCNN part is trained to maximise the log probability of the target pixels given the observed context pixels, while keeping the convolutional encoder and deconvolutional layers of the PixelVAE fixed. The objectives used in both stages are given in the next sections.
During test time, if the latent variables are inferred and target pixels sampled using the bidirectional PixelCNN, we obtain an ordinary semantic inpainting method. If the user manipulates the values of the latent variables directly, the generated content will match the high-level semantics given by the user-set values, while the low-level details are still consistent with the given context.

\section{Disentangled PixelVAE} \label{sec:disentangled_pixelvae}

\def\xx{\textbf{x}}
\def\xc{\textbf{x}_c}
\def\xt{\textbf{x}_t}
\def\zz{\textbf{z}}
\def\mm{\textbf{m}_c}
\def\EE{\mathbb{E}}
\def\loss{\mathcal{L}}
\def\reg{\mathcal{R}}
\def\KL{\mathsf{KL}}
\def\ptrain{p_\text{train}}
\def\MI{\mathsf{MI}}
\def\TC{\mathsf{TC}}
\def\PD{\mathsf{PD}}

In this section we will describe in detail the disentangled PixelVAE. We start with introducing some notation. We denote the image using $\xx$ and latent variables using $\zz$. We assume for simplicity that the image $\xx$ is $M\times M$, and the pixels are indexed in top-to-bottom, left-to-right raster scan order. The image is split into a context region $c$ and a target region $t$.  We use $\xc$ and $\xt$ to denote the pixel values for the context and target respectively. More specifically, $\xc = \mm \odot \xx$ and $\xt = (1-\mm)\odot \xx$, where  $\mm$ is the context mask, with value 1 for pixels in the context and 0 for those in the target. 

The disentangled PixelVAE component of our model consists of three modules. A convolutional encoder which maps the context $\xc$ to a distribution $q(\zz|\xc)$ over the latent variables $\zz$, a deconvolutional network which decodes $\zz$ to spatial feature maps, which is then transformed by the forward PixelCNN module into a distribution $p(\xx|\zz)$ over the whole image $\xx$ (the latter two forming the decoder). Our PixelVAE differs from the PixelVAE of \cite{gulrajani2016pixelvae} in three crucial aspects: the encoder is trained to work with masked images $\xc$ with widely different context regions, the latent variables are regularised to form a disentangled and informative representation of the image, and the spatial feature maps are fed into the PixelCNN in a slightly different way to accommodate the reverse PixelCNN module.  The rest of this section will describe the first two differences while the third is addressed in the next section.

\textbf{Encoding the context.} The aim of the PixelVAE component is to learn a latent variable representation of the high-level semantics of the whole image $\xx$, given only a partial observation of the context $\xc$. To this end, we train it using a stochastic auto-encoder loss of the form
\begin{align}
\loss(q,p) = \EE_{\ptrain(c)\ptrain(\xx)}[
\EE_{q(\zz|\xc,\mm)}[
-\log p(\xx|\zz)]] 
+ \reg(q)
\label{eq:objective1}
\end{align}
where $\ptrain(\xx)$ is the empirical distribution over training images, $\ptrain(c)$ is a training distribution over context regions for which we wish to train the encoder to handle, $q(\zz|\xc,\mm)$ is the stochastic encoder, $p(\xx|\zz)$ is the decoder, and $\reg(q)$ is a regulariser for the latent representation. The encoder is given only the context $\xc$, not the full image, to encourage it to learn to form a good representation of the whole image given only the observed context. It is also given access to the mask $\mm$ so that it knows which pixels are observed or unobserved. In the following, the latent variables are Gaussians, with means and diagonal variances parameterised by the convolutional encoder. Note that $\ptrain(c)$ is important and should be chosen to reflect the distribution over context regions at test time; we do not expect the encoder to do as well on context regions that are very different from those it is trained on.  In experiments we use a uniform distribution over a rectangle as the target region (the context consists of pixels outside the rectangle).

If the encoder uses the full image, $q(\zz|\xx)$, and the regulariser $\reg(q) = \EE_{p(\xx)}[\KL(q(\zz|\xx)\|p(\zz))]$ penalises the KL divergence between the variational posterior and some prior $p(\zz)$, then \eqref{eq:objective1} reduces to the standard VAE objective \citep{kingma2013auto,rezende2014stochastic}. If the encoder uses only the context, $q(\zz|\xc,\mm)$, as in \eqref{eq:objective1}, the objective still forms an upper bound on $-\log \int p(\xx|\zz)p(\zz)d\zz$, the negative log marginal probability of image $\xx$, albeit a looser one. However this looser objective can work better in the scenario we are interested in here, where only the context is observed, since it is trained accordingly.

\textbf{Regularising for a disentangled and informative latent representation.} In order that the latent variables can be used to control the semantic inpainting process, it is important that they form a representation of the image with two properties: that they are informative (so that different user-chosen values for latent variables lead to different inpainting results) and interpretable (so that users have an intuitive understanding of the effect of varying each latent variable). 
The choice of the regulariser is important in this respect, since it dictates the type of representation learnt.  This is currently a very active area of research, and we will provide here an short discussion of different regularisers in order to motivate our choice.  For simplicity we will assume that the encoder's input is the full image $\xx$ rather than the context $\xc$.

The starting point is the KL regulariser for the VAE, which has been decomposed into a number of distinct terms by \citet{makhzani2015adversarial,hoffman2016elbo,chen2018isolating} (assuming a factorial prior $p(\zz)=\prod_j p(\zz_j)$):
\begin{align}
\reg_\text{VAE}(q) =& \EE_{\ptrain(\xx)}[\KL(q(\zz|\xx)\|p(\zz))] 
\label{eq:decomposition} \\
=& 
\underbrace{
    \KL\infdivx{q(\zz,\xx)}{\ptrain(\xx)q(\zz)}
}_\text{
    Data-Latent Mutual Information (\textsf{MI})
} + 
\underbrace{
    \KL\large(q(\zz)\|\textstyle\prod_j q(\zz_j)\large)
}_\text{
    Total Correlation (\textsf{TC})
} + 
\underbrace{
    \textstyle\sum_j \KL\infdivx{q(\zz_j)}{p(\zz_j)}
}_\text{
    Divergence from prior ($\PD$)
}
\nonumber
\end{align}
where $q(\zz,\xx):=\ptrain(\xx)q(\zz|\xx)$ is the joint distribution of $\zz$ and $\xx$ defined by the training empirical distribution over $\xx$ and the encoder distribution of $\zz$ given $\xx$, $q(\zz)$ its marginal distribution over the latents, and $j$ ranges over the indices of the latent variables. 
Thus minimising $\reg_\text{VAE}$ tends to: \emph{minimise} mutual information between latent variables and data ($\MI$), make latent variables independent ($\TC$), and make latent variables marginally be close to the prior ($\PD$). 

By upweighting/downweighting each of these terms, latent representations with different properties can be learnt. 
For example, the $\beta$-VAE \citep{higgins2016beta} upweights all three terms, so leads to disentangling of latent variables due to higher $\TC$ penalty, but also blurry reconstructions due to low information content caused by higher $\MI$ penalty.
More recently, $\beta$-TCVAE \citep{chen2018isolating} and FactorVAE \citep{kim2018disentangling} addressed this by only penalizing $\TC$ more, leaving $\MI$ and $\PD$ unchanged.
Another line of research focused the so-called information preference problem \citep{makhzani2015adversarial,chen2016variational,zhao2017infovae}, whereby the latent variables tend to be ignored if the decoder can successfully model the data well by itself, which is the case here since we use a PixelCNN decoder. 
\citet{chen2016variational} proposed to limit the receptive field size of the decoder so that it will only be able to model local details, and the model is forced to use the latent variables to model global structure. 
InfoVAE \citep{zhao2017infovae} and adversarial autoencoder (AAE) \citep{makhzani2015adversarial} both propose to disregard the $\MI$ term in \eqref{eq:decomposition}, so that mutual information between inputs and latent representations is not penalized. 
In the appendix, Table \ref{table:training_objective_candidates} summarises the various alternative regularisers. 

Returning to our requirements for the latent representation, which are that they should be both informative and disentangled, there are three possible solutions. First, we can use $\beta$-TCVAE/FactorVAE to penalise the total correlation $\TC$ to encourage disentangling, but limiting the size of the PixelCNN receptive fields to disallow it from modelling global semantic properties. Second, we can exclude the mutual information term $\MI$ instead of limiting the receptive field size of the PixelCNN, and at the same time, further penalise the total correlation $\TC$ for disentangling; we call this Info-$\beta$-TCVAE. Finally, we can use InfoVAE-MMD, aka MMD-VAE, which does not penalise the mutual information $\MI$, but uses maximum-mean discrepancy (MMD) \citep{gretton2012kernel} to measure divergence between the aggregated posterior $q(\zz)$ and the prior $p(\zz)$ ($\approx$ \textsf{TC+PD}). By analogy with $\beta$-VAE, we believe scaling up the MMD regulariser should lead to good disentangling, with little negative consequence due to the absence of the mutual information term $\MI$.

In section \ref{sec:experiments}, we empirically compare among VAE, Info-$\beta$-TCVAE and InfoVAE-MMD, and find that heavily penalising total correlation $\TC$ also leads to the model ignoring the latent variables, which is previously believed to be the effect of penalising the mutual information $\MI$. The phenomenon was not observed in \citep{kim2018disentangling,chen2018isolating}, and we believe this is because they used a weaker decoder.
Moreover, we empirically discover that, under the asymmetric KL divergence, removing the mutual information penalty $\MI$ leads to low variance of the aggregated posterior $q(\zz)$ (which is supposed to match the prior $p(\zz)$). The phenomenon suggests that, it may be essential to use symmetric divergences such as maximum-mean discrepancy (as in \citet{chen2016variational}) or Jensen-Shannon divergence (as in \citet{makhzani2015adversarial}) if mutual information penalty $\MI$ is disregarded. We leave further discussion to Section \ref{sec:experiments}.
As a result, our regulariser of choice is the InfoVAE-MMD objective.

\section{Bidirectional PixelCNN}\label{sec:bidirectional_pixecnn}

In this section we describe the architecture and training regime of the bidirectional PixelCNN part of our model. Recall that we train our model in two stages: first we train the PixelVAE to learn a high-level disentangled semantic representation of the image, while the second stage trains the bidirectional PixelCNN, consisting of both a forward PixelCNN (shared with the PixelVAE) and reverse PixelCNN, to capture the low-level details given the high-level semantics. 
Recall that for ease of exposition we take the image $\xx$ to be $M\times M$. The high-level semantics is captured by the latent variables $\zz$, which are transformed by the deconvolutional layers of the PixelVAE into $M\times M$ spatial feature maps, which we denote using $Y^z$ in the following.

We start with a review of PixelCNNs
\citep{oord2016pixel,van2016conditional}, which model the joint distribution of image pixels as $p(\xx) = \prod_{k=1}^{M^2} p(x_k| x_1,\ldots,x_{k-1})$, where pixels are arranged in top-to-bottom, left-to-right raster scan order. The network architecture consists of $L$ convolutional layers, each of which takes as input a tensor of shape $(B,M,M,C_{l-1}^f)$ and outputs one of shape $(B,M,M,C_l^f)$, where $B$ is the batch size and $C_l^f$ are the number of channels at layer $l$. Given a batch of images $X$ as input, the network computes an output tensor $Y^f_L$ such that $Y^f_L[b,i,j,:]$ parameterises the conditional distribution of $X[b,i,j,:]$, the $(i,j)$th pixel of the $b$th image in the batch. This conditional distribution should depend only on the pixels preceding the $(i,j)$th one in the raster scan order, and this is achieved by carefully designed convolutional filters which mask out later pixels in the order. 
Typically, PixelCNNs are trained to optimise the log probability of images. Because of the raster scan order, when using the resulting trained PixelCNN for inpainting, each inpainted pixel will be consistent with pixels to its top and left, but not to the context pixels to its bottom and right, creating a jarring line to the bottom and right of the target region. 

We address this by using a second PixelCNN that operates in the reverse raster scan direction. This reverse PixelCNN takes as input $X_c$, the batch of masked context pixels along with the binary mask itself, and outputs $Y^r$, a tensor of shape $(B,M,M,C^r)$, which captures the dependence on context pixels below and to the right of each pixel.  The masking of convolutional filters in this reversed PixelCNN is like the forward one, except it is simply rotated $180^\circ$. 
$Y^r$, along with the spatial feature maps $Y^z$ capturing high-level semantics (a tensor of shape $(B,M,M,C^z)$ where $C^z$ is the number of channels in the feature maps), are then fed into the forward PixelCNN as location-dependent biases.  Specifically, for the $l$th layer of the forward PixelCNN, it has input $Y^f_{l-1}$ and output $Y^f_{l}$, which is computed as:
\begin{align}
    Y^f_l = \sigma(
    W^f_l * Y^f_{l-1} + 
    U_l * Y^r +
    V_l * Y^z
    )
    \label{eq:bidirectional_layer}
\end{align}
where $*$ denotes convolution, $W^f_l$ is the masked filters for the forward PixelCNN, $U_l$ and $V_l$ are $1\times 1$ unmasked convolutional filters, and $\sigma$ is the nonlinearity. For simplicity, we have omitted technical details such as gates \cite{van2016conditional} and resnet blocks \cite{salimans2017pixelcnn++}.

We train the bidirectional PixelCNN to optimise the loss
\begin{align}
    \EE_{\ptrain(c)\ptrain(\xx)}\left[
    \EE_{q(\zz|\xx_c)}\left[
    -\sum_{k\in t}\log p(\xx_k | \xx_{t[<k]},\xx_c, \zz)
    \right]\right] 
    \label{eq:bidirectional_loss}
\end{align}
where $k\in t$ ranges over the pixel indices in the target $t$, and $x_{t[<k]}$ denotes the set of pixels in the target preceding pixel $k$ in the forward raster scan order. The latent variables $\zz$ are drawn from the convolutional encoder distribution of the PixelVAE, and dependence on it is through the spatial feature maps $Y^z$ computed by the deconvolutional layers of the PixelVAE. Each term in \eqref{eq:bidirectional_loss} can be read off from a slice of the output of the forward PixelCNN, $Y^f_L[b,i,j,:]$, where $(i,j)$ corresponds to the image location of pixel $k$, and $b$ is the batch index.  The forward filters $W^f_l$ are initialised from the first stage training of the disentangled PixelVAE, while $U_l$, $V_l$ and the parameters of the reverse PixelCNN are initialised as per usual.  

For inpainting, the target pixels $\xx_t$ are sampled in the forward raster scan order, conditional on $\xx_c$. The method does not require post-processing like Poisson blending \citep{perez2003poisson} since the bidirectional PixelCNN ensures local consistency with all surrounding context pixels.  By ignoring the term $V_l * Y^z$ in \eqref{eq:bidirectional_layer}, the bidirectional PixelCNN (without PixelVAE) is an interesting semantic inpainting model in its own right. It can be used to generate a diversity of plausible inpainting results, but cannot be easily controllable by the user.

\section{Experiments} \label{sec:experiments}

We experimentally evaluate the proposed model and conduct ablation studies. Section \ref{subsec:semantic_inpainting} shows that the proposed model can generate target pixels which are coherent with the context pixels at both the high semantic level and the local detail level. In Section \ref{subsec:controllable_semantic_inpainting} we demonstrate that by varying latent variables, our method can generate plausible inpaintings matching user-specified high-level semantics. Section \ref{subsec:ablation_study} reports ablation studies which show the effects of choosing different training regimes, including with/out the reverse PixelCNN module and one/two-stage training.

We perform experiments on the CelebA dataset \citep{liu2015faceattributes}. The dataset contains $202259$ human face images with great diversity in terms of poses and expressions. To focus on the variation of human faces rather than the background, the images are cropped at the center to $128\times128$, and are then resized to $32\times32$, which is suitable for our available computing resources. The model is trained on $200000$ images, and is evaluated on the $2259$ held-out images. 
We use $32$ dimension latent variables for all experiments. For the network architecture and other details about implementation, please refer to Section \ref{sec:implementation_and_hyperparameters}.

\subsection{Uncontrolled Semantic Inpainting} \label{subsec:semantic_inpainting}

\begin{figure}
  \centering
    \includegraphics[width=0.8\linewidth]{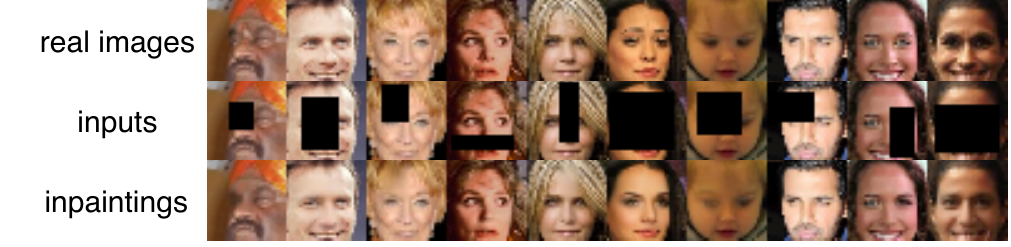}
\caption{Semantic inpainting on \emph{held-out} images with randomly sampled rectangle masks.}
\vspace*{-1em}
\label{fig:semantic_inpainting}
\end{figure}

We apply random rectangle masks to held-out images, and rely on our model to fill in the missing regions. Widths, heights and positions of the masks are sampled uniformly. Some real images, masked images, and inpainting results are shown in Figure \ref{fig:semantic_inpainting}. In this experiment, values of latent variables are directly sampled from the posterior. Our inpainting method generates content which is semantically coherent, and locally consistent with the given context without any post-processing. The inpainting results are plausible even when missing regions are very large.

\subsection{Controllable Semantic Inpainting} \label{subsec:controllable_semantic_inpainting}

\begin{figure}
  \centering
    \includegraphics[width=1\linewidth]{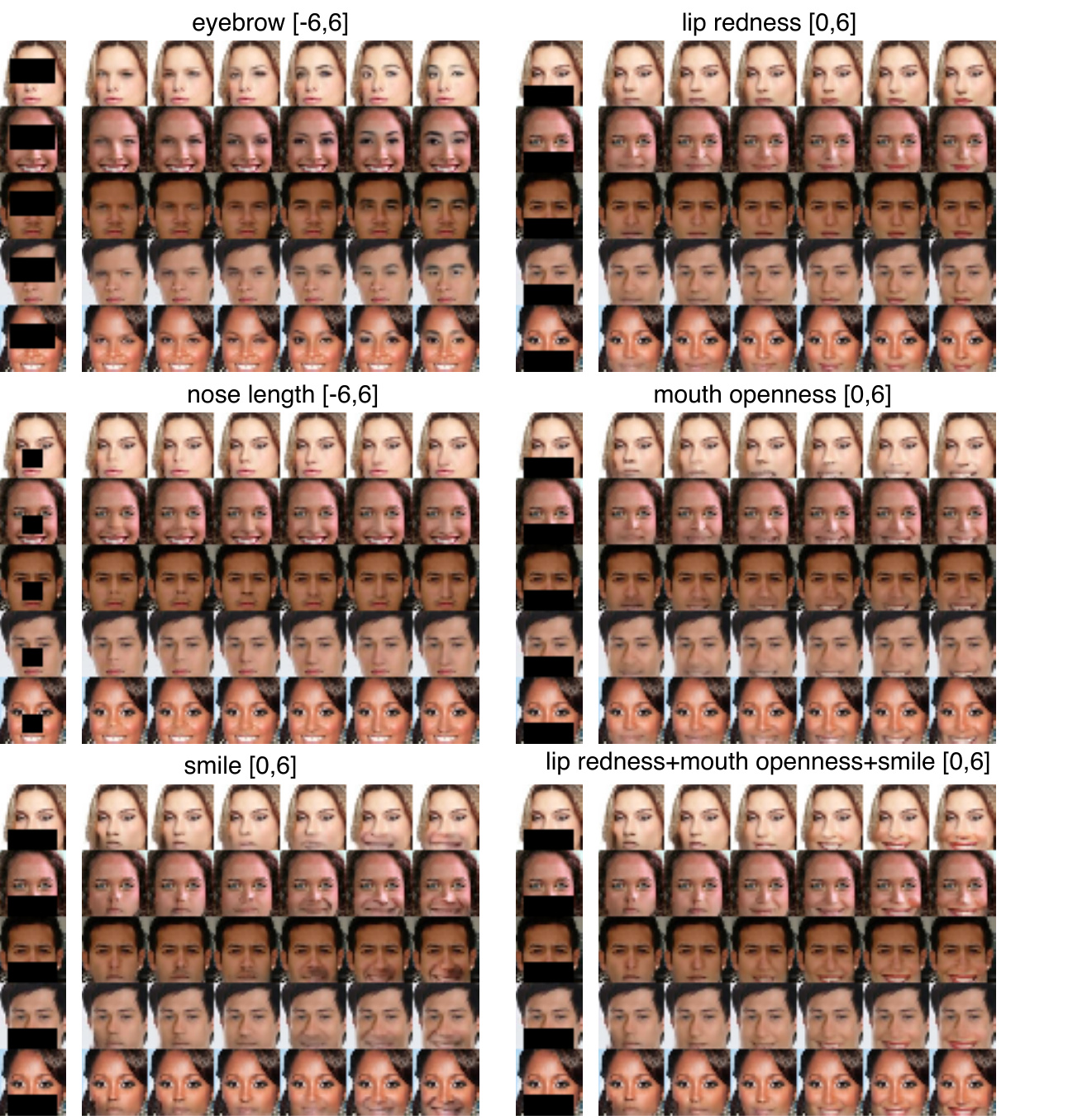}
\caption{User-controllable semantic inpainting on \emph{held-out} images. Latent variables corresponding to certain features are set by users during test-time, while other latent variables inferred. In order to show the effects of varying these user-set latent variables, values are set to be evenly spaced over the interval specified in the brackets, and corresponding inpainting results are tiled in the same row.}
\label{fig:CSI}
\end{figure}

We apply rectangle masks to several distinct areas of human faces, such as the eyes, the nose and the mouth, and vary latent variables associated with certain features of the masked regions. In Figure \ref{fig:CSI}, we demonstrate five independent factors, each of which is controlled by a single latent variable. Furthermore, we vary three factors relating to the mouth region together, and demonstrate that effects of these factors can be superimposed.

These interpretable factors can be discovered by latent traversal, where each time we vary only a single latent variable while keeping all the rest fixed to their inferred values. Many latent factors corresponds to global features such as gender, azimuth, skin tone, hue, brightness, etc. We did not try to vary them because it will lead to global inconsistency. 

\subsection{Ablation Study} \label{subsec:ablation_study}

\begin{figure}
\begin{subfigure}{0.49\textwidth}
\centering
\includegraphics[width=0.8\linewidth]{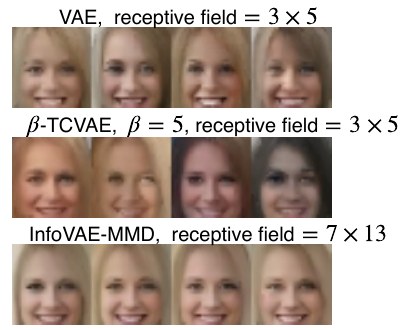}
\caption{Penalising total correlation heavily also leads to ignorance of latent code. Images on the same row are generated by multiple runs of reconstruction with the same model.}
\label{subfig:total_correlation}
\end{subfigure} \hfill
\begin{subfigure}{0.49\textwidth} 
\centering
\includegraphics[width=0.8\linewidth]{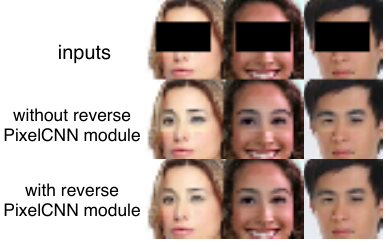}
\caption{Bidirectional PixelCNN. Discontinuities on the bottom and right edges of target regions can be observed in the second row, but do not occur in the bottom row.}
\label{subfig:bidirectional_pixelcnn}
\end{subfigure} \hfill
\begin{subfigure}{1\textwidth} 
\centering
\includegraphics[width=0.6\linewidth]{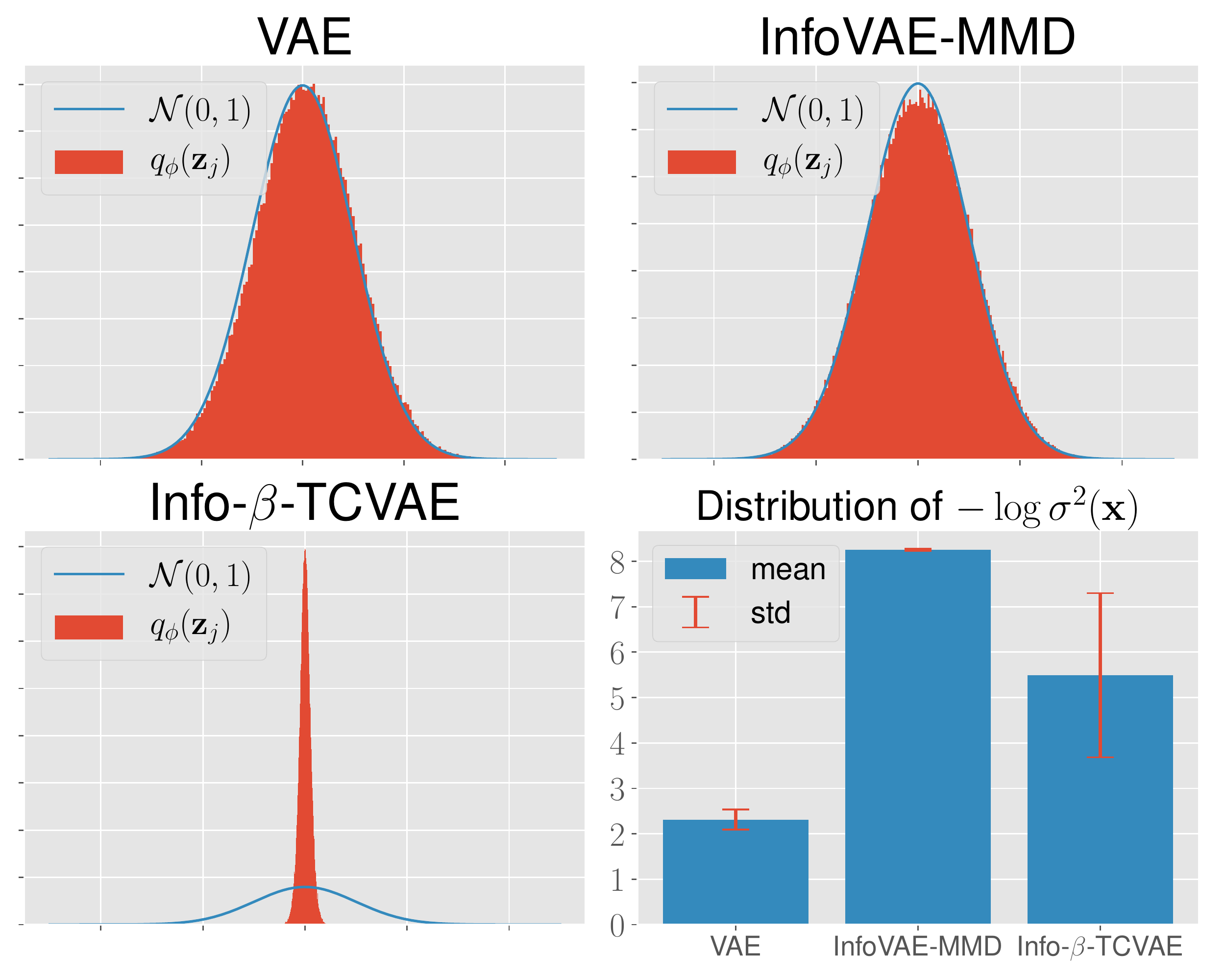}
\caption{Failed regularisation of Info-$\beta$-TCVAE. First three figures show marginal aggregated posterior (red) and the prior (blue). In the last figure $\sigma^2(\xx)=var[q(\zz_j | \xx)]$ where $\xx \sim \ptrain(\xx)$. Here we only show $j=0$, but similar phenomenon can be observed for other $j$}
\label{subfig:z_dist} 
\end{subfigure}
\caption{Ablation study}
\end{figure}

\textbf{Training objectives.} 
We train the disentangled PixelVAE architecture with the VAE, the $\beta$-TCVAE ($\beta=5$), and the InfoVAE-MMD objectives. We use a small receptive field $3\times5$ for the first two, and a big receptive field $7\times15$ for the last. Here a receptive field of $3\times5$ means, in the PixelCNN module, each pixel has access to pixels in the $3\times5$ rectangle area above, and $\frac{5-1}{2}$ pixels to the left of it. We feed the same test image into models and run reconstruction $4$ times for each model. The results are shown in Figure \ref{subfig:total_correlation}. If latent variables capture most high-level semantics, each model should produce very similar reconstruction images with minor difference in details. Otherwise, latent variables learnt may be uninformative, and data distribution is mostly modelled by the powerful decoder. Figure \ref{subfig:total_correlation} shows that, even with a small receptive field, the model with $\beta$-TCVAE objective produces very diversified samples, showing serious ignorance of latent code. Note that the only difference between VAE and $\beta$-TCVAE here is a heavy penalty of total correlation. That is to say, heavily penalising total correlation $\TC$ alone makes the information preference problem serious. The phenomenon is understandable, because in the extreme case where latent code is completely random isotropic Gaussian noise, total correlation decreases to zero. Generating uninformative latent code seems to be easier than learning informative disentangled latent representations. 

On the other hand, a natural development to overcome the problem of $\beta$-TCVAE is to still penalise total correlation $\TC$, but at the same time, encourage data-latent mutual information $\MI$. Info-$\beta$-TCVAE ($=\beta\TC+\PD$) is such a candidate. However, in Figure \ref{subfig:z_dist}, we show this is also not an option, at least under the KL divergence. As shown, density mass of the aggregated posterior gathers around the high probability region of the prior when Info-$\beta$-TCVAE is used. We believe the reason is the following: under the marginal KL divergence $\KL\infdivx{q(\zz)}{p(\zz)}$ ($=\TC+\PD$), $q(\zz)$ always tends to underestimate the support of $p(\zz)$. However, when the mutual information penalty $\MI$ is present, the posterior $q(\zz|\xx)$ is encouraged to have big variance. At the same time, as mentioned in \citet{zhao2017infovae}, the reconstruction term in \eqref{eq:objective1} (the likelihood term excluding the regulariser) always attempts to learn disjoint supports for posteriors with different observations, so that it can overfit the training set. Combing the reconstruction term and the mutual information term $\MI$, the posteriors with different observations are forced to spread out in order to make room for each other, thereby counter-balancing the effect of the marginal KL. However, when the mutual information term $\MI$ is absent, the counter-balancing force disappears.
We decide on the InfoVAE-MMD objective because it does not suffer from the latent code ignorance problem, and disentangled interpretable latent factors can be discovered by raising this regulariser.

\textbf{Two-stage training.} It is possible to train the model in one stage. The architecture is the same, and the training objective is the second-stage objective in \eqref{eq:bidirectional_loss} plus the regularisation term in \eqref{eq:objective1}. In practice, we find two problems when the model is trained in one stage (see Appendix \ref{sec:additional_experimental_results}). Firstly, latent factors have less control over the final outputs. Secondly, glaring noise around the bottom/right corner may appear if latent code is set by the user. For the first problem, just as stated in Section \ref{sec:a_controllable_semantic_inpainting_method}, we believe it is because our model extract more context information from the bidrectional PixelCNN stream rather than the disentangled PixelCNN stream. For the second problem, we conjecture that it is because the latent variables learn to encode mask information since it is only required to reconstruct the target pixels. Varying them then confuses the bidirectional PixelCNN.

\textbf{Bidirectional PixelCNN.} Finally, we remove the reverse PixelCNN module in our architecture so our model becomes a disentangled PixelVAE model trained in one stage. Recall that the forward PixelCNN decoder only directly model the dependence on context pixels above and to the left of each pixel, while dependence on bottom/right context pixels is modelled indirectly through the latent variables. Unsurprisingly, as shown in Figure \ref{subfig:bidirectional_pixelcnn}, when this model is used for controllable semantic inpainting, a line can be observed on the bottom/right boundary of the target region.
\FloatBarrier

\section{Conclusion and Future Work}

We present a controllable semantic inpainting method, where the user can interact with the inpainting process by specifying high-level semantics of the target regions. 
We enhance the conditional PixelCNN with a reverse PixelCNN to form the bidirectional PixelCNN, which we use to capture low-level details and texture in the context. 
We discuss the difficulty of finding training objectives when expressive decoders and disentangled latent representations are both desired under the VAE framework. We empirically and analytically compare different training objectives and explain our preferences. 
We also demonstrate that two-stage training and the bidirectional PixelCNN modules are important in our model.

Our results suggest several future directions. 
Firstly, learning disentangled latent factors and using an expressive decoder are both essential for our method, but combining the two makes the design of training objectives challenging. We expect there exist better solutions to this problem, and a further investigation is worthwhile.
Secondly, our latent factors are learnt in a purely unsupervised manner. But for some problems, we are able to provide supervision or semi-supervision, which should make latent factors even more interpretable. It would be promising to incorporate supervision into our model \citep{kingma2014semi,maaloe2016auxiliary,sohn2015learning,siddharth2017learning,bouchacourt2017multi}.

\newpage 
\begin{appendices}
\normalsize
\section{Implementation and Hyperparameters} \label{sec:implementation_and_hyperparameters}

\subsection{Network Architecture}

Architecture of our convolutional neural network and deconvolutional neural network modules is shown in table \ref{tb:network_architecture}. Note that the deconvolutional neural network outputs feature maps with $32$ channels rather than $3$ channel RGB images. In practice, missing regions of inputs are filled with uniform random noise rather than the special value $0$, to prevent feature extraction from these regions.

\begin{table}[!htb]
  \caption{Network architecture}
  \label{tb:network_architecture}
  \centering
  \small
  \begin{tabular}{|l|l|}
    \toprule
    \textbf{Convolutional neural network} & \textbf{Deconvolutional neural network} \\
    \midrule
    Input $32\!\times\!32\!\times\!4$ RGB Image $+$ Mask & Input $\zz \in \mathcal{R}^{32}$  \\  \cmidrule{1-2}
    $1\!\times\!1$ conv $32$. stride $1$. \textit{SAME}. batchnorm. ELU      & FC $512$. batchnorm. ELU \\ \cmidrule{1-2}
    $4\!\times\!4$ conv $64$. stride $2$. \textit{SAME}. batchnorm. ELU      & $4\!\times\!4$ deconv $256$. stride $1$. \textit{VALID}. batchnorm. ELU \\ \cmidrule{1-2}
    $4\!\times\!4$ conv $128$. stride $2$. \textit{SAME}. batchnorm. ELU      & $4\!\times\!4$ deconv $128$. stride $2$. \textit{SAME}. batchnorm. ELU \\ \cmidrule{1-2}
    $4\!\times\!4$ conv $256$. stride $2$. \textit{SAME}. batchnorm. ELU      & $4\!\times\!4$ deconv $64$. stride $2$. \textit{SAME}. batchnorm. ELU \\ \cmidrule{1-2}
    $4\!\times\!4$ conv $512$. stride $1$. \textit{VALID}. batchnorm. ELU      & $4\!\times\!4$ deconv $32$. stride $2$. \textit{SAME}. batchnorm. ELU \\ \cmidrule{1-2}
    FC $32$. batchnorm. None  \big|   FC $32$. batchnorm. None   & \\ \cmidrule{1-2}
    \bottomrule
  \end{tabular}
\end{table}

Our implementation of the PixelCNN module and the reverse PixelCNN module is based on \citetapp{salimans2017pixelcnn++} and their released code. For both modules, we use $100$ filters for all layers except for the last one. Unless otherwise stated, we use a receptive field of $7\times15$, which means we stack $5$ gated Resnet blocks (Please refer to the public repository of PixelCNN++ for more details, and our code will be released later.) We use a dropout \citepapp{srivastava2014dropout} rate of $0.5$ during training for all PixelCNN layers. However, we do not use downsampling or dilated convolution to model long range structure, because in our model, only details are handled by PixelCNNs. We use exponential linear units (ELU) \citepapp{clevert2015fast} as activation functions in all modules. 

\subsection{Other Hyperparameters}

We globally initialise all parameters with Xavier initialisation \citepapp{glorot2010understanding}. For stochastic optimisation, we use ADAM \citepapp{kingma2014adam} with an initial learning rate of $0.0001$ and a batch size of $64$. We use a large coefficient ($2\times10^6$) to scale up the MMD regulariser, and it leads to the discovery of many interpretable latent factors without sacrificing reconstruction quality. To model colour channels of a pixel, we use discrete logistic mixture with $10$ components (see \citetapp{salimans2017pixelcnn++}). In terms of sampling from the distribution parameterised by the PixelCNN outputs, it has previously been discovered that, controlling the concentration of sampling distribution, where pixel values are actually sampled from a tempered softmax, leads to significantly better visual quality \citepapp{dahl2017pixel,kolesnikov2017pixelcnn}. We apply a similar idea but with slightly different implementation which is more efficient under the discrete logistic mixture distribution. We sample colour values from truncated components in range $[\mu_i-s_i,\mu_i+s_i]$, where $\mu_i$ and $s_i$ are the mean and the scale of the $i$th mixture component.

\subsection{Implementation}

Our model was implemented in Tensorflow \citepapp{abadi2016tensorflow}. We adopt the implementation of maximum-mean discrepancy (MMD) \citepapp{gretton2012kernel} at \href{https://ermongroup.github.io/blog/a-tutorial-on-mmd-variational-autoencoders/}{https://ermongroup.github.io/blog/a-tutorial-on-mmd-variational-autoencoders/}, with exactly the same hyperparameters and the radial basis function kernel.

\section{Training Objectives} \label{sec:training_objectives}

For clarity, we provide a comparison of different training objectives in Table \ref{table:training_objective_candidates}.

\begin{table}[!htb]
  \centering
  \small
  \begin{threeparttable}[b]
  \caption{Training objective candidates}
  \label{table:training_objective_candidates}
  \begin{tabular}{lll}
    \toprule
    Objectives     & Regularization     & Divergence \\
    \midrule
    VAE \citep{kingma2013auto} & $\textsf{MI}+\textsf{TC}+\textsf{PD}$  & KL     \\
    $\beta$-VAE \citep{higgins2016beta} & $\beta(\textsf{MI}+\textsf{TC}+\textsf{PD})$ & KL \\
    $\beta$-TCVAE \citep{chen2018isolating},
    FactorVAE \citep{kim2018disentangling} 
    & $\textsf{MI}+\beta\textsf{TC}+\textsf{PD}$ & KL\\
    InfoVAE-MMD \citep{zhao2017infovae}, 
    AAE \citep{makhzani2015adversarial}     
    & $\approx \textsf{TC+PD}$ & MMD, GAN  \\
    Info-$\beta$-TCVAE & $\beta\textsf{TC}+\textsf{PD}$ & KL \\
    \bottomrule
  \end{tabular}
  \end{threeparttable}
\end{table}

To evaluate these training objectives, we need to compute mutual information $\MI$, total correlation $\TC$, and divergence from prior $\PD$ separately. Therefore, we need to compute entropy of the aggregated posterior $\EE_{q_{\phi}(\zz)}[-\log q_{\phi}(\zz)]$ along with its dimension-wise counterpart, where $q_{\phi}(\zz)=\EE_{\ptrain(\xx)}[q_{\phi}(\zz\mid \xx)]$. Exact computation is intractable, so we estimate the negative entropy term using a mini-batch importance sampling estimator proposed by \citetapp{chen2018isolating,esmaeili2018hierarchical}:
\begin{equation}
    \begin{aligned}
    \EE_{q_{\phi}(\zz)}\Big[\log q_{\phi}(\zz)\Big] \approx \frac{1}{M} \sum_{m=1}^{M} \log \Big[\frac{1}{N}\Big[ q_{\phi}(\zz\mid \xx_m) + \frac{N-1}{M-1} \sum_{m^{\prime}\neq m} q_{\phi}(\zz\mid \xx_{m^{\prime}})\Big] \Big],
    \end{aligned}
\label{eq:is_estimator}
\end{equation}
where $\{\xx_1,...,\xx_M\}$ is a random batch and $N$ is the overall dataset size. The estimator is biased due to concavity of the logarithm, but the bias is small enough when the batch size is sufficiently large \citep{esmaeili2018hierarchical}.

\section{Additional Experimental Results} \label{sec:additional_experimental_results}

\subsection{Visualisation of Latent Factors}

\begin{figure}
    \centering
  \includegraphics[width=0.8\linewidth]{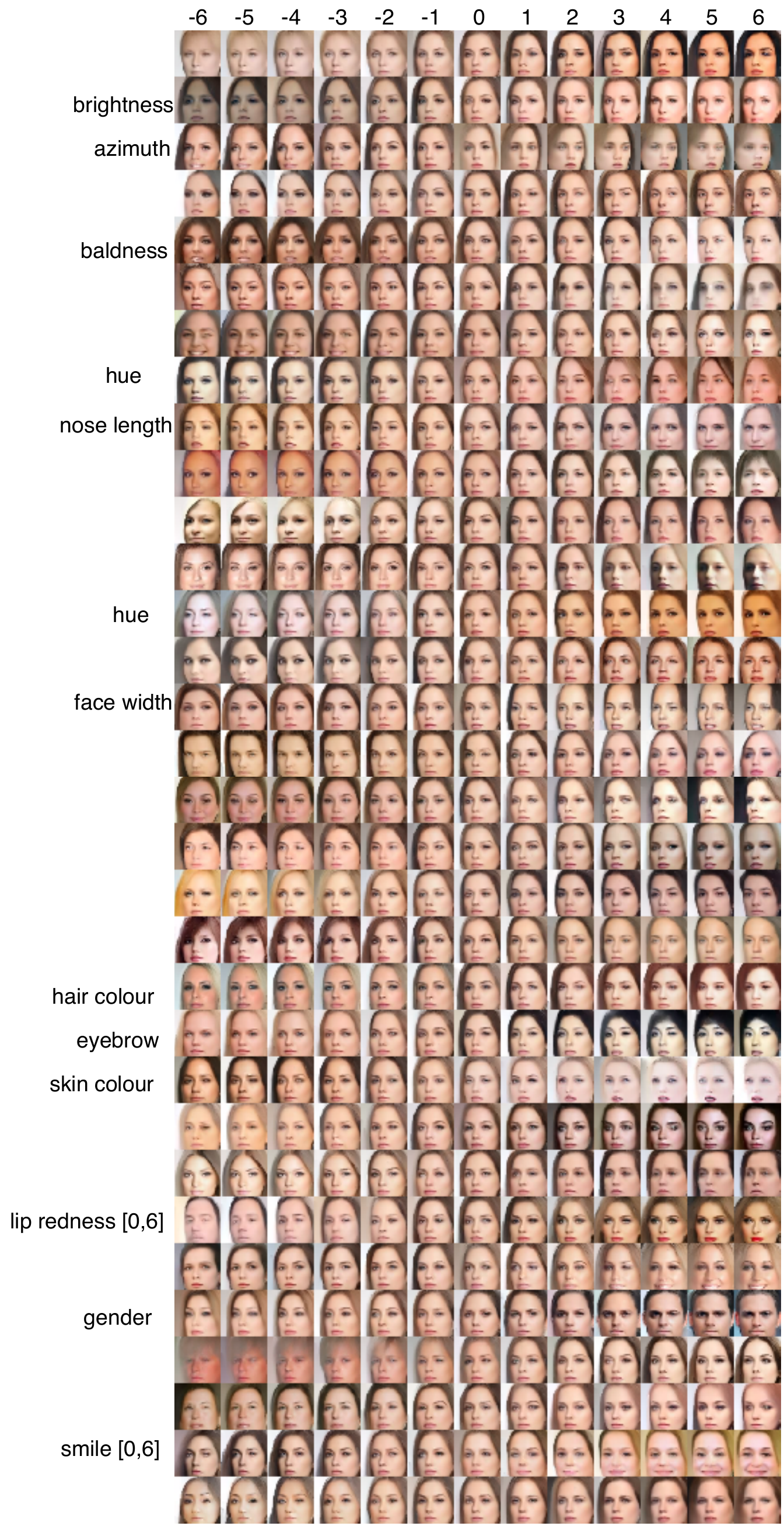}
  \caption{Visualisation of latent factors. We vary a single latent variable each time (see values at the top), and keep other variables fixed to their inferred values.}\label{fig:latent_traversal}
\end{figure}

To visualise latent factors, we feed a seed image into the network, vary a single latent variable each time while keeping all other variables fixed to their inferred values, and then reconstruct the whole image without the reverse PixelCNN module. We show this latent code traversal in Figure \ref{fig:latent_traversal}, where each row corresponds to a latent variable, and the user-set values are shown at the top of the figure. 

We discover several interpretable factors as shown in the figure. Note that some factors are only manifested in one direction of the latent variables. In terms of disentangling, the performance is not as good as those reported in \citetapp{chen2018isolating,kim2018disentangling}, where a weak decoder is used and the total correlation is heavily penalised. However, the powerful autoregressive decoder is critical in our model, and the InfoVAE-MMD objective can handle the latent code ignorance problem very well. So we still decide on this objective.

\newpage 
\subsection{Varying Global Factors}

\begin{figure}
    \centering
  \includegraphics[width=0.8\linewidth]{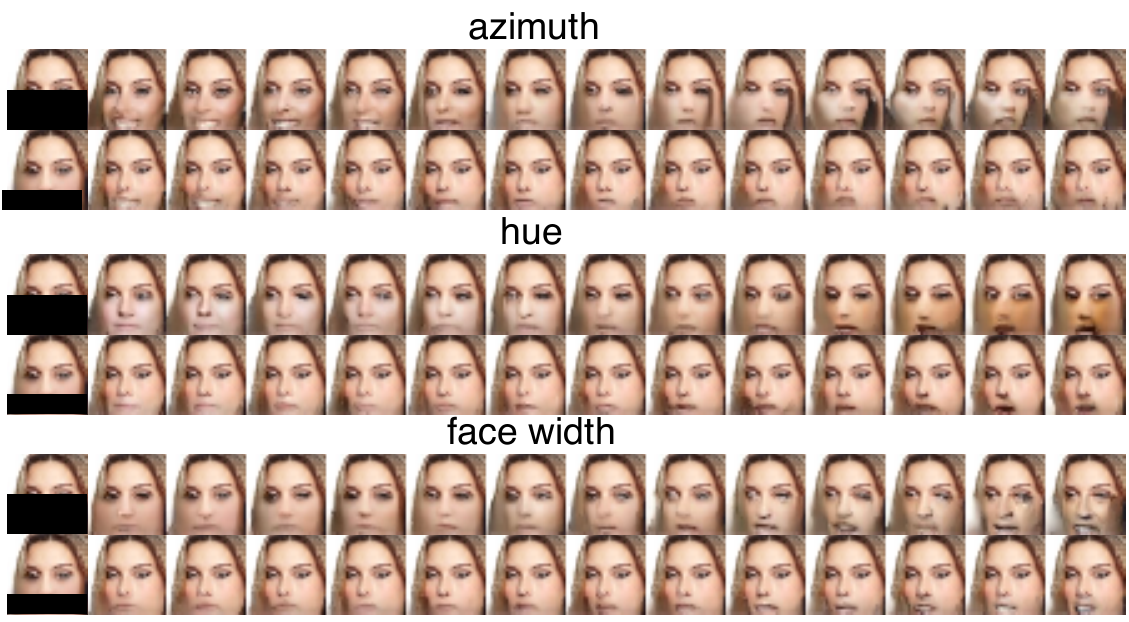}
  \caption{Varying global factors. User-set values are evenly spaced over $[-6,6]$.}\label{fig:global_factors}
\end{figure}

Some latent factors correspond to global features, such as hue, gender, face width, etc. Figure \ref{fig:global_factors} shows effects of varying these global factors.

Generally speaking, when missing regions are big enough, the model will try very hard to match these global features. On the other hand, when missing regions are small, the model tends to ignore latent factors and rely more on context information extracted from the bidirectional PixelCNN stream. In Figure \ref{fig:global_factors} you can see that, when bottom half of the test image is masked, the model still tries to show variations of azimuth, hue and face width, even though the results may lose coherence. when bottom $1/4$ is masked, we cannot observe corresponding effects of these factors on the inpaintings.

\subsection{Two-stage Training}

\begin{figure}
    \centering
  \includegraphics[width=0.45\linewidth]{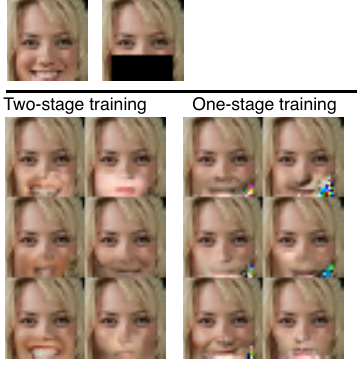}
  \caption{Ablation study: two-stage training. Instead of sampling from the actual posterior, all latent variables are directly sampled from $\mathcal{N}(\mathbf{0},3\mathbf{I})$ (so high-level context semantics is completely abandoned). The model trained in two stages produces inpainting results with more diversity than the model trained in one stage. In addition, we can sometimes observe jarring noise on the bottom-right corner when one-stage training is used. }\label{fig:two_stage}
\end{figure}

To demonstrate why two-stage training is essential, we also train our model in one stage. The architecture is the same, and the training objective is the second-stage objective in \eqref{eq:bidirectional_loss} plus the regularisation term in \eqref{eq:objective1}. 
With a demonstrative inpainting task shown in Figure \ref{fig:two_stage}, we want to know, to what extent can latent variables control the inpainting results. So instead of using inferred values, we sample all latent variables directly from $\mathcal{N}(\mathbf{0},3\mathbf{I})$, and show inpainting results in Figure \ref{fig:two_stage}. We run experiments $6$ times for both models (the model trained in one stage and the model trained two-stage). 
The model with two-stage training generates more diversified inpaintings than the model trained in one stage. Roughly speaking, diversity here suggests more overall control over the inpainting process by latent code. This observation accords with our intuition that two-stage training forces the model to extract context information as much as possible from the disentangled PixelCNN stream rather than the bidirectional PixelCNN stream, so that latent variables will take more responsibility. In addition, glaring noise in the target regions can sometimes be observed if the model is trained in one stage. This phenomenon is not observed if latent representations are inferred. Therefore, our conjecture is, since the model only needs to reconstruct target pixels rather than complete images in the one-stage training, latent variables may try to only form representations of the target pixels, making latent representations mask-dependent. Varying them makes the reverse PixelCNN hard to distinguish between real context pixel values and masked pixel values (which are set to $0$).

\end{appendices}
\end{document}